\documentclass[letterpaper, 10 pt, conference]{ieeeconf}  

\IEEEoverridecommandlockouts                              

\usepackage{graphics} 
\usepackage{epsfig} 
\usepackage{mathptmx} 
\usepackage{times} 
\usepackage{amsmath} 
\usepackage{amssymb}  
\usepackage[normalem]{ulem}
\usepackage{hyperref}
\title{\LARGE \bf
Score refinement for confidence-based 3D multi-object tracking
}

\author{Nuri Benbarka,$^{1}$ Jona Schr\"oder,$^{1}$ and  Andreas Zell$^{1}$
\thanks{$^{1}$Nuri Benbarka, Jona Schr\"oder, and  Andreas Zell are with the cognitive systems group,
        University of T\"ubingen, Sand 1, T\"ubingen, Germany
        {\tt\small first-name.last-name@uni-tuebingen.de}}}%

\begin{document}

\maketitle
\thispagestyle{empty}
\pagestyle{empty}

\begin{abstract}
	
   Multi-object tracking is a critical component in autonomous navigation, as it provides valuable information for decision-making. Many researchers tackled the 3D multi-object tracking task by filtering out the frame-by-frame 3D detections; however, their focus was mainly on finding useful features or proper matching metrics. Our work focuses on a neglected part of the tracking system: score refinement and tracklet termination. We show that manipulating the scores depending on time consistency while terminating the tracklets depending on the tracklet score improves tracking results. We do this by increasing the matched tracklets' score with score update functions and decreasing the unmatched tracklets' score. Compared to count-based methods, our method consistently produces better AMOTA and MOTA scores when utilizing various detectors and filtering algorithms on different datasets. The improvements in AMOTA score went up to 1.83 and 2.96 in MOTA. We also used our method as a late-fusion ensembling method, and it performed better than voting-based ensemble methods by a solid margin. It achieved an AMOTA score of 67.6 on nuScenes test evaluation, which is comparable to other state-of-the-art trackers. Code is publicly available at: \url{https://github.com/cogsys-tuebingen/CBMOT}.

\end{abstract}

\section{INTRODUCTION}

3D multi-object tracking (MOT) aims to find the objects surrounding an agent in 3D space and trace them through time. The trajectories built by the MOT algorithms are used by motion forecasting modules or given directly to the motion planner to complete navigation successfully. There is a tendency to use online MOT algorithms \cite{yin2020center,weng2019baseline,chiu2020probabilistic} due to their simplicity, speed, and accuracy. These algorithms filter outliers in frame-by-frame object detectors \cite{lang2019pointpillars,shi2019pv,cheng2020improving} by utilizing temporal information. Also, because they rely heavily on object detection performance, the notable advancement in 3D object detection resulted in considerable MOT growth.

In online MOT, deciding when to initialize and terminate a tracklet is critical to MOT performance. In the previous works \cite{weng2019baseline, chiu2020probabilistic}, this decision was \textit{count-based}. In those methods, The tacklet's output is only considered if it has a minimum number of consecutive detections (\textit{min-hits}). On the other hand, a tracklet is terminated, if it is not updated for a predefined number of timesteps (\textit{max-age}). \textit{Min-hits} generally lowers false positives and increases false negatives,  and \textit{max-age} enhances robustness against missed detections and occlusions but increases false positives.

The \textit{count-based} method's problem is that it treats all detections the same, yet the detection score can indicate the detection quality. For example, if the detection score is 0.95, it is probably a true-positive. In this case, why should the algorithm wait for more matches if it is already confident about the first detection? Moreover, if the tracklet's initial score is low, why should it remain for many timesteps before it is terminated? Furthermore, when a detection matches a tracklet, the tracklet's updated score is the detection score. Here the tracklet's previous time step score is not used, and we believe it is a valuable information source that can improve the score estimation. For these reasons, we use a \textit{confidence-based} method for initialization and termination. 

The \textit{confidence-based} method initializes a tracklet and considers its output when its score is higher than the detection threshold (\textit{det-th}). And it terminates the tracklet when its score goes below the deletion threshold (\textit{dlt-th}). Moreover, the tracklet's score decreases in the estimation step by a constant value (\textit{score-decay}) and, if it is matched with a detection, it increases by the \textit{score update function}. In this case, tracklets consistently matched over time will have high scores, and unmatched tracklets' scores will decline. To the best of our knowledge, \cite{sun2019scalability} was the only work that used \textit{confidence-based tracking}. They added the detection's and tracklet's scores to update the tracklet's score. Our work will show that their \textit{score update function} performance is poor, and in the best cases, it will work like the \textit{count-based} method. Consequently, we will show that \textit{confidence-based} MOT outperforms \textit{count-based} MOT if we employ proper \textit{score update functions}.

We propose various score update functions. 
Our functions lead to more stable tracking scores, which eventually lead to better performance. We tested our method on both nuScenes \cite{caesar2020nuScenes} and Waymo \cite{sun2019scalability} datasets with multiple detectors and filtering algorithms. Results showed that the \textit{confidence-based} method with score refinement consistently provides higher AMOTA and MOTA results than \textit{count-based} methods with our proposed functions.

\begin{figure*}
	\begin{center}
		\includegraphics[width=0.8\textwidth]{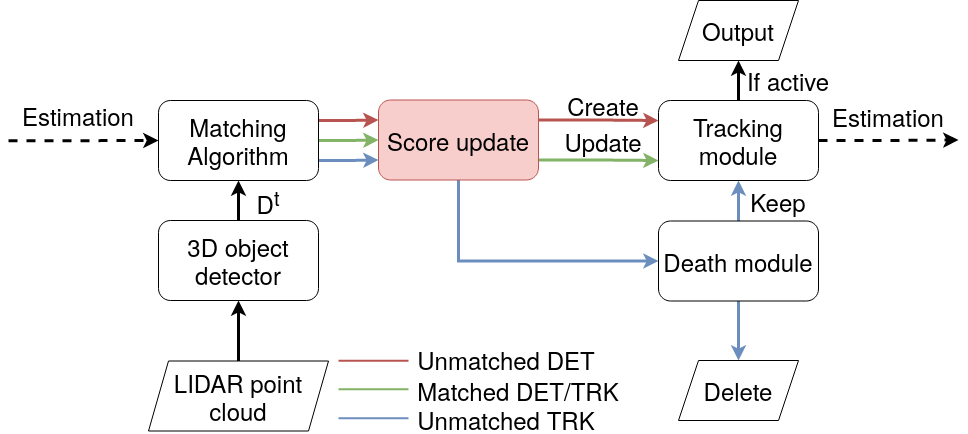}
	\end{center}
	\caption{Our pipeline consists of five main modules: 1) 3D object detector, 2) Matching Algorithm, 3) Score update, 4) Filtering algorithm, and 5) Object death module. DET and TRK are short forms of detections and tracklets.}
	\label{fig:pipeline}
\end{figure*}

\section{RELATED WORK}

\subsection{3D object detection}

A 3D MOT system's performance is mainly affected by object detection performance \cite{weng2019baseline, weng2020gnn3dmot}. A 3D object detector processes point clouds and gives back 3D bounding boxes for every frame as input to 3D multi-object tracking systems. One of the first algorithms trained end-to-end on point clouds for the 3D detection task was VoxelNet \cite{zhou2018voxelnet}. It used PointNets \cite{qi2017pointnet} to produce point features inside each voxel. These features are processed with 3D sparse convolutions, then passed into a region proposal network. And finally, it uses 2D convolutions to produce the detections. SECOND \cite{yan2018second} sped up VoxelNet by using spatially sparse convolutional networks to extract features from the z-axis before the 3D data is downsampled to 2D features. More techniques attempted to eliminate the costly 3D convolutions, PIXOR \cite{yang2018pixor} projected all points onto a 2D feature map with 3D occupancy and point intensity data encoded in the feature dimension. PointPillars \cite{lang2019pointpillars} used pillars instead of voxels to give pillar features; it used 2D convolutions on the pillar feature, which provided an efficient backbone. 

PointRCNN \cite{shi2019pointrcnn} took another approach and avoided voxelization, and directly operates on 3D point clouds. It generates proposals and then refines them to generate the detections using PointNets. PV-RCNN \cite{shi2020pv} merged the voxel-based and point-based pipelines to get the advantages of both. Other works focused more on discovering helpful features for detection. For example, MVF \cite{zhou2020end} fused features from bird's-eye view (BEV) and perspective views of the same lidar point cloud and introduced the concept of dynamic voxelization, where there is no need to set a fixed amount of points per voxel. PointPainting \cite{vora2020pointpainting} added semantic segmentation information to point clouds as visual features. Hu et al. \cite{hu2020you} used free space as additional information to the detector. CenterPoint \cite{yin2020center} considered object detection as a keypoint estimation problem as was done in \cite{zhou2019objects} for 2D detection. It predicts a heatmap for each of the classes and then uses the heatmap's local minima features to regress to 3D bounding boxes. 

\subsection{3D Multi-Object Tracking}

In general, MOT is divided in terms of data association into batch and online tracking; batch MOT aims to find the global data association by finding the minimum cost of a flow graph \cite{schulter2017deep}. In online MOT or sometimes called tracking-by-detection, the goal is to perform data association with the previous frame only, where it becomes a bipartite matching problem. It is usually solved using the Hungarian algorithm \cite{kuhn1955hungarian}. For 3D MOT, online MOT became popular nowadays because of its simplicity, efficiency, and influence of the 2D MOT methods \cite{bergmann2019tracking,bewley2016simple,wojke2017simple,zhou2020tracking}.

AB3DMOT \cite{weng2019baseline} built a simple pipeline where they used a Kalman filter \cite{kalman1960new} for tracking objects and the Hungarian algorithm for data association between the tracklets and detections. They used IoU as a matching metric and showed a great performance in terms of speed and accuracy. Chiu et al. \cite{chiu2020probabilistic} used the Mahalanobis distance \cite{mahalanobis1936generalized} instead of the IoU. They also found the Kalman filter parameters automatically from the statistics of the detector's training results. FANTrack \cite{baser2019fantrack} integrated 2D appearance features from CNNs and 3D bounding box features for data association. GNN3DMOT \cite{weng2020gnn3dmot} proposed using four features: motion and appearance from 2D and 3D spaces. And used graph neural networks to learn the interaction between these features.

However, all these works focused on feature selection or matching-metric selection. They were using count-based tracking for initializing and terminating tracklets and didn't look into score updating. The closest work to ours is Sun et. al \cite{sun2019scalability}, who used confidence-based tracking and employed addition as their update function. However, adding the scores makes the system extra-confident and results in high numbers of false positives. We propose more proper \textit{score update functions} and show that \textit{confidence-based} MOT performs better than count-based MOT.

\section{METHOD}

Our 3D tracking pipeline has the structure shown in figure \ref{fig:pipeline}. It can work on any 3D detector such as VoxelNet \cite{zhou2018voxelnet} or Pointpillars \cite{lang2019pointpillars} and any filtering algorithm, such as the Kalman filter \cite{weng2019baseline}\cite{chiu2020probabilistic}, extended Kalman filter, or the point tracker used in CenterPoint \cite{yin2020center}. In the following sections, we will describe the modules in the pipeline.

\subsection{3D object detector}
\label{3d_detector}
A 3D object detector processes point clouds and returns 3D bounding boxes. At each time step $ t $, the detector produces a number $ n $ of detections $D^t = \{ D_1^t, D_2^t, \dots, D_n^t \}$. Each detection $ D_i^t $ is a tuple of 8 parameters $ (x,y,z,l,w,h,\theta,s) $. Here, the parameters $ (x,y,z) $ are the object's center in either the vehicle's local frame or the global frame, the parameters $ (l,w,h) $ are the bounding box dimensions, $\theta$ is the yaw angle, and $ s $ is the detection score.

\subsection{Matching module}

This module is responsible for matching the detected 3D bounding boxes and the tracking module estimations. There are mainly two algorithms for this task; the Hungarian algorithm or a greedy algorithm \cite{chiu2020probabilistic,yin2020center}. Both need a metric to perform the matching, and examples of the metric used are IoU \cite{weng2019baseline}, euclidean distance \cite{yin2020center}, Mahalanobis distance \cite{chiu2020probabilistic}, and distance in embedding space \cite{weng2020gnn3dmot,baser2019fantrack}.

\subsection{Tracking module}
The tracking module is responsible for four tasks: to give new tracklets an ID and save them in memory, decide whether a tracklet should be \textit{active} or not, estimate the tracklet's parameters in the next time step, and update the tracklet's information according to its matched detection using a filtering algorithm (e.g. a Kalman filter).

The tracklets' state depends on the filtering algorithm used. However, it should have at least ten parameters $ (x, y, z, l, w, h, \theta, id, c, active) $. The first seven parameters are the first seven parameters of the detection tuple discussed in section \ref{3d_detector}. $ id $ is an identifier that is unique for every instance in the sequence. $ c $ is the tracklet score which tells how much the system is confident about the tracklet. And $ active $ is a Boolean variable stating whether the tracklet is active or not. When a tracklet is not \textit{active}, the tracking module does not consider it an output yet keeps it in memory. The module sets the tracklet's \textit{active} variable to true if it is matched with a detection or it is newly created and has a score higher than the \textit{detection threshold}. Also, it sets the \textit{active} variable of unmatched tracklets kept by the object death module to false if their score is lower than the \textit{active threshold}.

\begin{figure}
	\begin{center}
		\includegraphics[width=0.5 \textwidth]{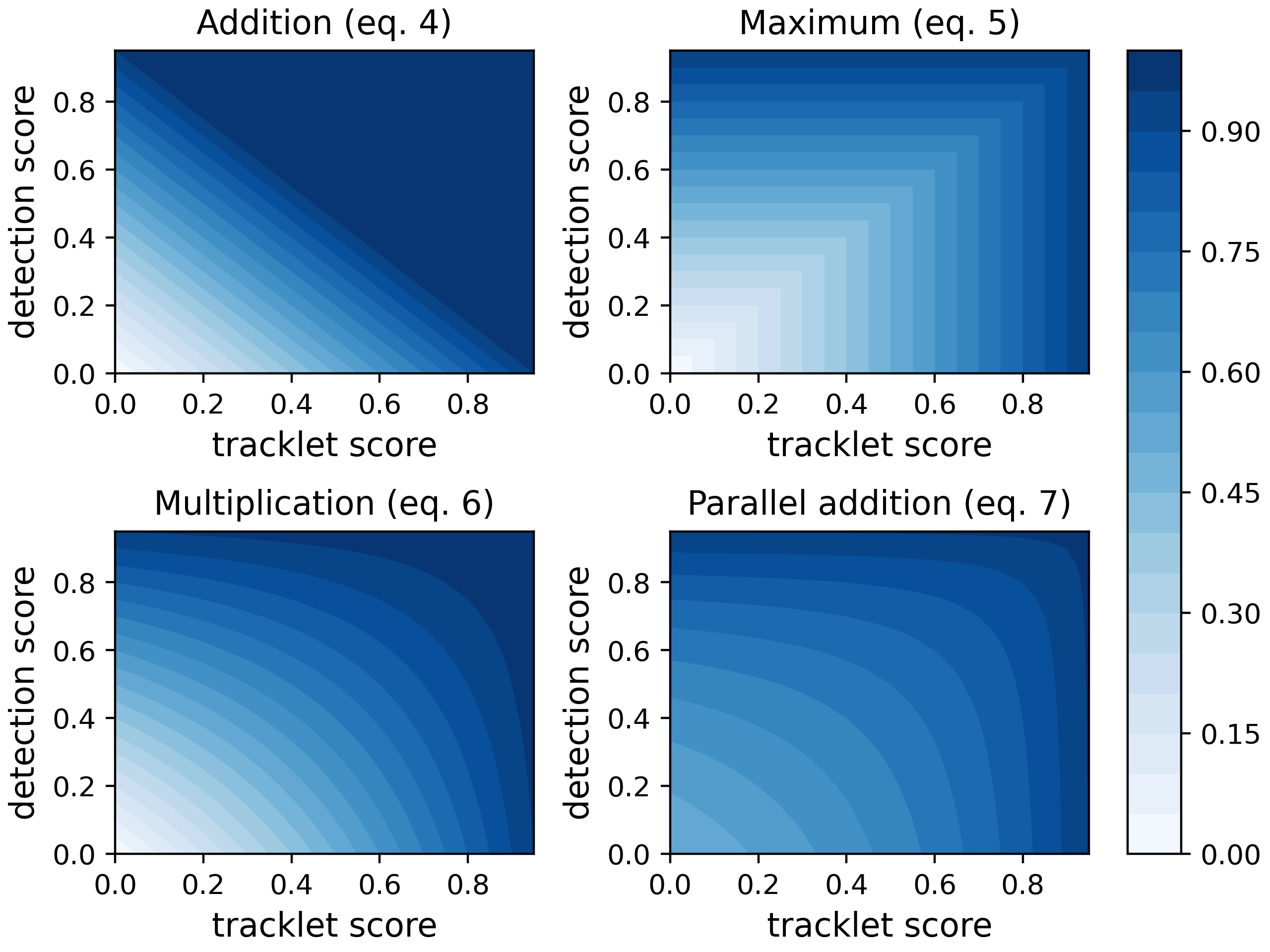}
	\end{center}
	\caption{Contour plots of the different update function outputs.}
	\label{fig:score_update}
\end{figure}

\subsection{Score-update}
This is the main module in our work, and it is responsible for manipulating the tracklet's scores. In \textit{count-based} methods, the tracklet score is simply the score of the matched detection, as shown in equation \ref{eq:confdet}
\begin{equation}
\label{eq:confdet}
c_{t} = s_t.
\end{equation}
$ c_t $ is the tracklet score, and $ s_t $ is the matched detection's score of the current time step $ t $. And if there is no matched detection, the tracklet score is unchanged $ c_{t} = c_{t-1} $. We think that this approach has two disadvantages. First, we believe that the algorithm should believe less in the tracklet presence by time unless continuously matched with detections. For this reason, the score update module reduces the tracklets' score by a constant \textit{score-decay} $ \sigma_{score} $  to get an estimated score of the current time step, as shown in equation \ref{eq:confpr}.

\begin{equation}
\label{eq:confpr}
\hat{c}_{t} = c_{t-1} - \sigma_{score}
\end{equation}    
here $ \hat{c}_{t} $ is the estimated tracklet score of the current time step. The second disadvantage of the \textit{count-based} method is that the new tracklet score does not depend on the previous time step's score $ c_{t-1} $ if there is a match. We believe that $ c_{t-1} $ is a valuable information source that can better measure the current score $ {c}_{t} $. Therefore, we use update functions that are dependant on the previous tracklet and detection scores. Furthermore, we argue that if there is a match, the algorithm should be more confident about its decision than its detection's confidence or the previous time step's confidence. Accordingly, we set the criterion for selecting an update function to give a score greater than or equal to both the tracklet and detection scores, as shown in equation \ref{eq:cri}.
\begin{equation}
\label{eq:cri}
c_t = f(\hat{c}_{t}, s_t) \geq max(\hat{c}_{t}, s_t)
\end{equation}
One function that satisfies this criterion is adding the tracklet and detection scores (equation \ref{eq:confadd}), as was used by \cite{sun2019scalability}. 
\begin{equation}
\label{eq:confadd}
c_{t} = \hat{c}_{t}+ s_t
\end{equation}
We reason that adding the scores will make the algorithm overconfident, and as seen in figure \ref{fig:score_update}, it is quite saturated. To tackle this problem of overconfidence, we propose the functions given in equations \ref{eq:confmax}, \ref{eq:confupmul}, and \ref{eq:confuppar}, which satisfy the above criterion and provide more robust scores.

\begin{align}
c_{t} &= max(\hat{c}_{t}, s_t) \label{eq:confmax}\\
c_{t} &= 1 - ((1 - \hat{c}_{t}) \cdot (1 - s_t)) \label{eq:confupmul}\\
c_{t} &= 1 - \frac{(1 - \hat{c}_{t}) \cdot (1 - s_t)}{(1 - \hat{c}_{t}) + (1 - s_t)} \label{eq:confuppar}
\end{align}

For equation \ref{eq:confmax}, it is the minimum requirement to satisfy the criterion above. And as for equations \ref{eq:confupmul} and \ref{eq:confuppar}, they are seeking to decrease the scores' complements. The intuition for that is if the score indicates the detection's confidence, then the score's complement indicates its uncertainty. If we multiply (eq. \ref{eq:confupmul}) or parallel add (eq. \ref{eq:confuppar}) these uncertainties, they become smaller, so naturally, the confidence gets higher. These equations give a middle ground between adding the scores and taking the scores' maximum, as shown in figure \ref{fig:score_update}. Finally, the new score of unmatched tracklets is the estimated score $ c_t = \hat{c}_t $.

\subsection{Object death module}
\label{birth}
All unmatched tracklets from the matching module are sent to the death module to decide whether to keep or delete the tracklet. The decision can be \textit{count-based}, \textit{confidence-based} or a mix of both. In \textit{count-based}, the module deletes the tracklet if it was unmatched for a number (\textit{max-age}) of timesteps. In \textit{confidence-based}, the module deletes the tracklet if its score goes below the \textit{deletion threshold}. We can use a combination of both; however, only one will dominate the performance. 

\begin{figure}
	\begin{center}
		\includegraphics[trim={35.5cm 0cm 0cm 0cm},clip,width=0.4\textwidth]{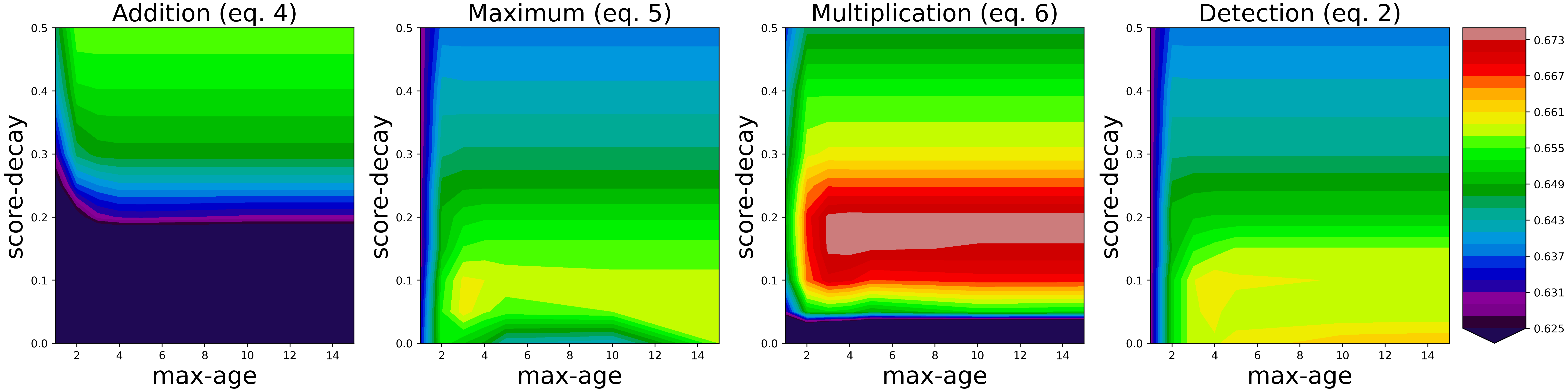}
	\end{center}
	\caption{Contour plots of the AMOTA results for the \textit{score-decay} experiment.}
	\label{fig:exp_detection}
\end{figure}

\section{EXPERIMENTS}

We did our tests on the nuScenes dataset \cite{caesar2020nuScenes}, a dataset used for autonomous driving tasks benchmarking. It contains 1000 driving scenes, each with a duration of 20 seconds captured in Boston and Singapore, and has 23 object classes annotated with 3D bounding boxes at a rate of 2Hz. The primary metrics used for evaluating MOT are Multi-Object Tracking Accuracy (\textit{MOTA}) and average \textit{MOTA} (\textit{AMOTA}). \textit{MOTA} is expressed as:
\begin{equation}
\textit{MOTA} = 1 - \frac{FP + FN + IDS}{GT}
\end{equation}
$ FP $ is the number of false positives, $ FN $ is the number of false negatives, $ IDS $ is the number of identity switches, and $ GT $ is the number of ground truths. \textit{MOTA} is found after fine-tuning the thresholds for each class, and then the maximum value is reported. On the other hand, \textit{AMOTA} measures the average performance of different thresholds, and it is expressed as: 
\begin{equation}
\textit{AMOTA} = \small \frac{1}{n-1} \sum_{r \in \{\frac{1}{n-1}, \frac{2}{n-1} \, ... \, \, 1\}} {\textit{MOTAR}}
\end{equation}
where $ r $ is the recall value, $ n $ is the number of recall values to test on, and \textit{MOTAR} is the recall-normalized \textit{MOTA}, and it is given as:
\begin{equation}
\begin{aligned}
\max \left(0,1 \, - \, \frac{{IDS}_r + {FP}_r + {FN}_r + (1-r) \cdot {GT}}{r \cdot {GT}}\right)
\end{aligned}
\end{equation} 
here $ {IDS}_r $, $ {FP}_r $, and $ {FN}_r $ are the id switches, false positives, and false negatives at a specific recall value $ r $.

\begin{figure}
	\begin{center}
		\includegraphics[trim={35.5cm 0cm 0cm 0cm},clip,width=0.4\textwidth]{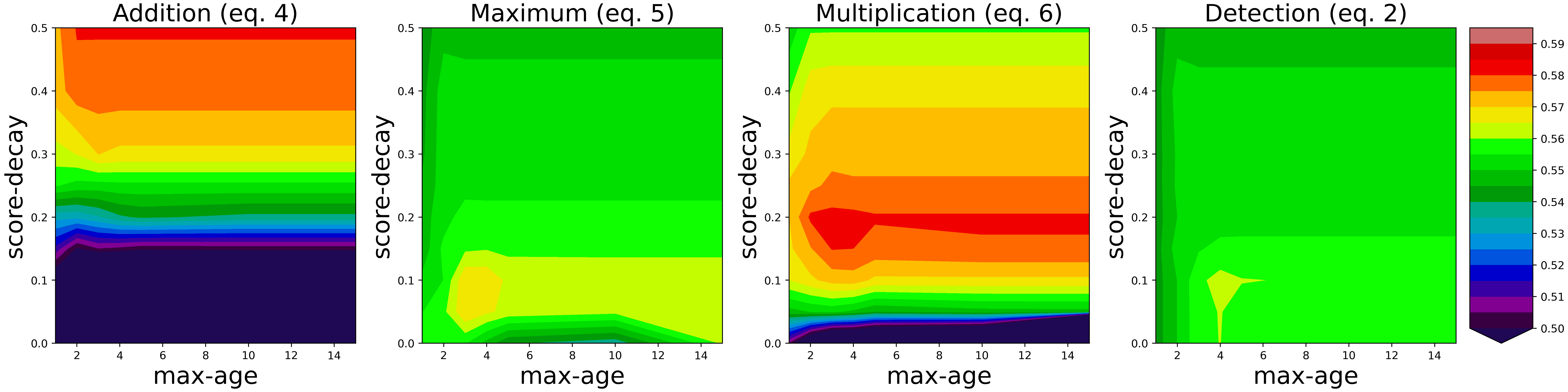}
	\end{center}
	\caption{Contour plots of the MOTA results for the \textit{score-decay} experiment..}
	\label{fig:mota_detection}
\end{figure}
\subsection{Ablation study}
\label{ssec:Ablation}
We use CenterPoint \cite{yin2020center} as a 3D detector and its point tracker as the filtering algorithm. We choose the greedy algorithm as the matching algorithm and the euclidian distance as the matching metric. We began with the following hyperparameters:  \textit{max-age} = 3, \textit{score-decay} = 0.0, \textit{deletion threshold} = 0.0 , \textit{active threshold} = 1.0, \textit{detection threshold} = 0.0, \textit{min-hits} = 1 and the \textit{update function} was equation \ref{eq:confdet}. This configuration produced the best results as reported in CenterPoint, and we made it our starting point for the ablation study. The next sections will discuss the impact of each hyperparameter.

\subsubsection{Score-decay and max-age}
\label{ssec:score-max}
We performed a grid search with the values of (0, 0.05, 0.1, 0.15, 0.2, 0.25, 0.3, 0,4, 0.5) for \textit{score-decay} and (1, 2, 3, 4, 5, 15) for \textit{max-age}, and figures \ref{fig:exp_detection} and  \ref{fig:mota_detection} show a contour of the AMOTA and MOTA results, respectively. We observe that the system gives high AMOTA and MOTA results at low \textit{score-decay} and large \textit{max-age}. We also see when we increase the \textit{score-decay} or lower the \textit{max-age}, the performance deteriorates.

The reason is that when we use equation \ref{eq:confdet} as a score updating function, the detection score overwrites the tracklet score. Thus, changing the \textit{score-decay} essentially affects the tracklet's lifetime, which is similar to changing \textit{max-age}. However, there is a slight difference between \textit{score-decay} and \textit{max-age} impacts; tracklets with low scores are deleted faster with \textit{score-decay} than with \textit{max-age}. This difference gave a 0.2 MOTA improvement at a \textit{score-decay} of 0.1; however, this improvement is not significant.

\begin{figure*}[h]
	\centering
	\includegraphics[width=\textwidth]{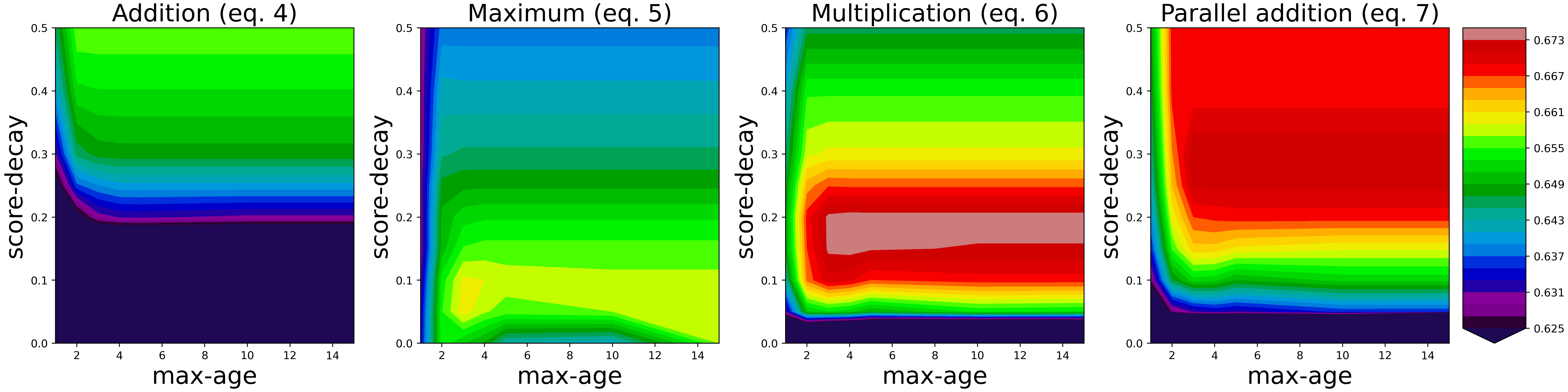}
	\caption{Contour plots of the AMOTA results of the different score update functions.}
	\label{fig:amota_noise}
\end{figure*}
\begin{figure*}[h]
	\centering
	\includegraphics[width=\textwidth]{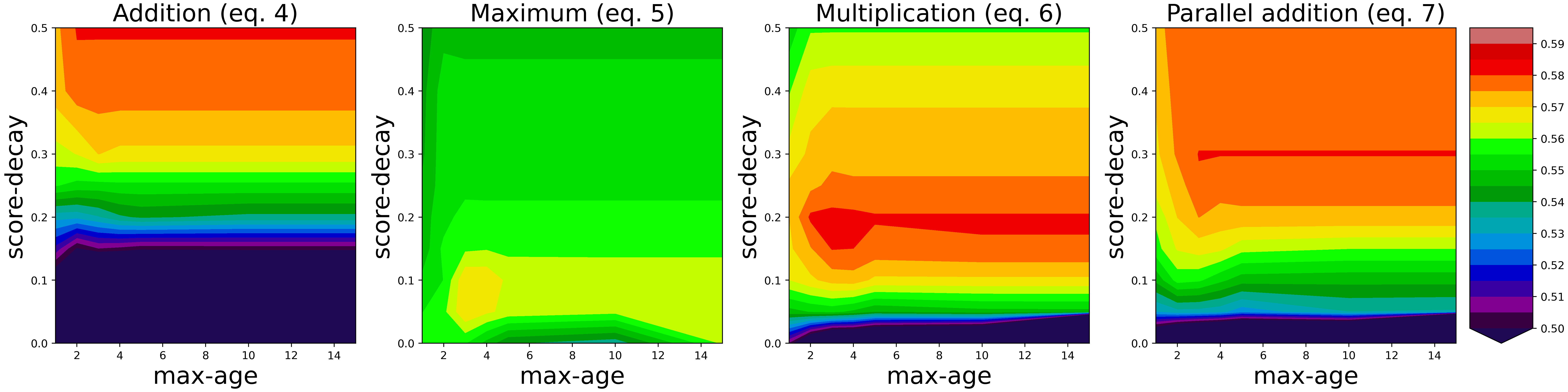}
	\caption{Contour plots of the MOTA results of the different score update functions.}
	\label{fig:mota_noise}
\end{figure*}

\subsubsection{Score update}
We repeated the experiment in section \ref{ssec:score-max} but with the other score-update functions (equations \ref{eq:confadd}, \ref{eq:confmax}, \ref{eq:confupmul}, \ref{eq:confuppar}). Figures \ref{fig:amota_noise} and \ref{fig:mota_noise} show the AMOTA and MOTA results, respectively. We notice that for equations \ref{eq:confmax}, \ref{eq:confupmul}, and \ref{eq:confuppar}, when we increase the \textit{max-age}, the performance improves until it peaks at a value of 4 and remains constant. As for equation \ref{eq:confadd}, the performance is dependent on \textit{max-age} even at high values. However, we also see that it performs worse than the other functions, and we can infer that it is not a proper function for the task. Moreover, we can conclude from these results that \textit{max-age} can be neglected.

We also observe that when the \textit{score-decay} value is approaching zero, the performance is terrible, and this is more predominant in equations \ref{eq:confadd} and \ref{eq:confuppar}, which increase the score a lot in one update. From this observation, we can conclude that using update functions without the \textit{score-decay} does not work, and therefore, they should be applied together. 

Furthermore, we perceive that the contour's shape is an inverted parabola in the \textit{score-decay} direction for all update functions. However, the difference between the functions is the value and position of the inverted parabola's peak. The more the update function can increase the score in one update step, the higher the \textit{score-decay} it needs to peak. And in the extreme case of equation \ref{eq:confadd}, it needs a \textit{score-decay} of 0.5 to peak. This means that it will most likely floor the estimated tracklet score to zero before adding it to the detection score. In this case, the algorithm works similarly to \textit{count-based} methods, where the detection score overwrites the estimated tracklets score, making it a poor update function to choose.

\begin{figure*}[h]
	\centering
	\includegraphics[width=\textwidth]{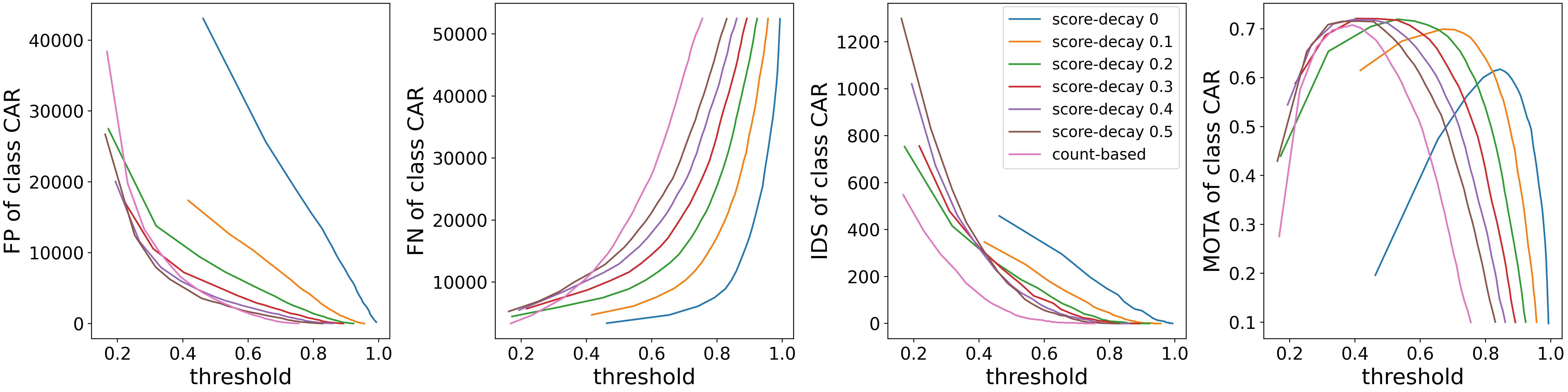}
	\caption{Metric details of class CAR vs score for different score-decay values.}
	\label{fig:noise_effect}
\end{figure*}

To further investigate what happens when changing the \textit{score-decay}, we plotted the metric details of class 'CAR' when using equation \ref{eq:confupmul} as an update function in figure \ref{fig:noise_effect}. We use the initial hyperparameters mentioned in section \ref{ssec:Ablation} for the baseline. We observe from figure \ref{fig:noise_effect} that when we increase the \textit{score-decay}, the whole FN curve shifts to the left, which results in shifting the MOTA peak to the higher thresholds and lowering its value. On the other hand, increasing the \textit{score-decay} reduces the FP and IDS curves' slope, which results in increasing the MOTA peak value and shifting its position to lower thresholds. We see from the MOTA curve that its peak is moving to the lower thresholds, and the peak value increases. This means that the \textit{score-decay} affects FP and IDS more than FN, which explains why our method works. In the rest of the ablation study, we pick equation \ref{eq:confupmul} as the score update function, set max-age to infinity (or not use it), and use a simple line search to find \textit{score-decay}.

\subsubsection{The rest of the hyperparameters}
We tested the rest of the hyperparameters, and we found that increasing \textit{min-hits} or the \textit{deletion-threshold} always reduces the performance. As for the \textit{active-threshold}, we found that decreasing it to 0.75 improved AMOTA by 0.03. And for the \textit{detection-threshold}, we found that increasing it to 0.15 further improved AMOTA by 0.05. We can conclude from the ablation study that our method has only one main hyperparameter, which is the \textit{score-decay}, and we can find its optimal value easily with a simple line search. The rest of the hyperparameters are either not needed (\textit{max-age}, \textit{min-hits}, and \textit{deletion-threshold}) or do not dramatically affect the performance (\textit{active-threshold} and \textit{detection-threshold}).
 
\begin{table*}[t]
	
	\begin{center}
		\begin{tabular}{c|c|c|c|c|c|c|c}
			\hline
			Detector & Tracker & SR &AMOTA$ \uparrow$ & MOTA$ \uparrow$  & FP$ \downarrow$ & FN$ \downarrow$ & IDS$ \downarrow$\\
			\hline
			CenterPoint& PointTracker& - &65.88  & 56.01  & \textbf{12295} & 21546 & \textbf{479}\\
			CenterPoint  &PointTracker& \checkmark &\textbf{67.51}  &\textbf{58.3}	& 13666 & \textbf{18882} & 494\\			
			\hline
			CenterPoint& KalmanFilter& - &65.39  & 55.33  & 13307 & 20559 & 811\\
			CenterPoint& KalmanFilter&\checkmark  &\textbf{67.22}  & \textbf{58.29}  & \textbf{13072} & \textbf{19534} & \textbf{610}\\
			\hline
			CBGS& PointTracker& - &60.13  & 51.82  & \textbf{10729} & 24116 & \textbf{754}\\
			CBGS& PointTracker&\checkmark  &\textbf{61.66}  & \textbf{54.20}  & 11408 & \textbf{22518} & 764\\
			\hline
			\hline
			CenterPoint*& PointTracker& - &63.84  & 53.66  & 18612 & \textbf{22928} & 760\\
			CenterPoint*&PointTracker& \checkmark &\textbf{64.93}  & \textbf{54.51}	 & \textbf{16469} & 24092 & \textbf{557}\\
			\hline
		\end{tabular}
	\end{center}
	\caption{Summary of the tracking results on the nuScenes dataset. SR is short for Score Refinement. *evaluated on the test split}
	\label{tab:summary_n}
\end{table*}
\begin{table*}[t]
	
	\begin{center}
		\begin{tabular}{c|c|c|c|c|c|c}
			\hline
			Detector & Tracker & SR & MOTA$ \uparrow$  & FP$ \downarrow$ & FN$ \downarrow$ & IDS$ \downarrow$\\
			\hline
			CenterPoint& PointTracker &- & 48.21  & 9.57 & \textbf{41.06} & 1.16\\
			CenterPoint & PointTracker& \checkmark  &\textbf{48.53}	& \textbf{9.41} & 41.31 & \textbf{0.75}\\
			\hline			
			CenterPoint& KalmanFilter& -& 50.15	 & \textbf{9.75} & 39.91 & 0.19\\
			CenterPoint& KalmanFilter & \checkmark & \textbf{50.76}  & 9.87 & \textbf{39.25} & \textbf{0.12}\\
			\hline
			PointPilllars & KalmanFilter & - & 40.51	& \textbf{9.69} & 49.66 & 0.14\\			
			PointPilllars & KalmanFilter & \checkmark& \textbf{41.01}& 10.19& \textbf{48.7}& \textbf{0.1}\\
			\hline
		\end{tabular}
	\end{center}
	\caption{Summary of the tracking results on the Waymo dataset. SR is short for Score Refinement.}
	\label{tab:summary_w}
\end{table*}

\begin{table*}[t]
	\begin{center}
		\begin{tabular}{c|c|c|c|c|c|c|c}
			\hline
			Ensemble method & LIDAR & Camera & AMOTA$ \uparrow$ & MOTA$ \uparrow$  & FP$ \downarrow$ & FN$ \downarrow$ & IDS$ \downarrow$\\
			\hline
			No fusion & \checkmark & - & 67.51 & \textbf{58.3} & 13666 & \textbf{18882} & 494\\
			No fusion & - & \checkmark & 17.75 & 15.04 & 15783 & 55009 & 7576\\
			\hline
			Affirmative & \checkmark & \checkmark & 67.35 & 52.65 & 18445 & 23194 & 360 \\
			Consensus/Unanimous & \checkmark & \checkmark & 50.80 & 48.73 & \textbf{8980} & 31573 & 659 \\ 
			\hline
			Confidence based & \checkmark $ \sigma_{score} $ & \checkmark & 67.33 & 52.86 & 18060 & 25485 & \textbf{353} \\
			 & \checkmark & \checkmark $ \sigma_{score} $ & 68.37 & 54.27 & 17874 & 20265 & 442 \\
		     & \checkmark $ \sigma_{score} $ & \checkmark $ \sigma_{score} $ & 68.73 & 54.48 & 15988 & 23235 & 529 \\
			 & \checkmark $ \sigma_{score} $* & \checkmark $ \sigma_{score}* $ & \textbf{69.18} & 56.30 & 17661 & 20304 & 459 \\
			
			\hline
		\end{tabular}
	\end{center}
	\caption{Tracking results with different ensemble methods. \quad $ \sigma_{score} $ Score-decay applied \quad * only if tracklets are unmatched}
	\label{tab:fusion}
\end{table*}

\begin{table*}[t]
	\begin{center}
		\begin{tabular}{c|c|c|c|c|c|c|c}
			\hline
			Method & LIDAR & Camera & AMOTA$ \uparrow$ & MOTA$ \uparrow$  & FP$ \downarrow$ & FN$ \downarrow$ & IDS$ \downarrow$\\
			\hline
		    AlphaTrack* & \checkmark & - & \textbf{69.3} & \textbf{57.6} & 18421 & 22996 & 718\\
			Octopus Tracker* & \checkmark & - & 67.9 & 57.2 & 16970 & \textbf{22272} & 781\\
			EagerMot\cite{kimeagermot} & \checkmark & \checkmark & 67.7 & 56.8 & 17705 & 24925 & 1156 \\
			\textbf{CBMOT} (ours) & \checkmark & \checkmark & 67.6 & 53.9 & 21604 & 22828 & 709 \\ 
			MCMOT* & \checkmark & \checkmark & 66.6 & 55.6 & \textbf{16322} & 23065 & 1803 \\
			Tracker* & \checkmark & - & 65.6 & 54.3 & 16631 & 24116 & 732 \\
		    CenterPoint\cite{yin2020center} & \checkmark  & - & 65.0 & 54.5 & 16469 & 24092 & \textbf{557} \\
			\hline
		\end{tabular}
	\end{center}
	\caption{
	Comparative evaluation on nuScenes test split. \quad * unpublished work}
	\label{tab:leaderboard}
\end{table*}

\subsection{Method generalization}

After we finished the ablation study, we wanted to check if our method generalizes well. We took the best configuration from the ablation study and applied it to different filtering algorithms and detectors. We used CBGS \cite{zhu2019classbalanced} as the second detector and Kalman Filtering as a second filtering algorithm. As for the Kalman filter, its state is a 6D vector consisting of the center position, velocity, and acceleration in the \textit{x} and \textit{y} directions. 

Tables \ref{tab:summary_n} and  \ref{tab:summary_w} show that score refinement always improves the results even without further tuning the hyperparameters. We also uploaded our tracking results on the test split on the evaluation server, and we got an improvement of 1.1 in AMOTA against CenterPoint \cite{yin2020center}. We say that these improvements are consistent and computationally free, so there is no reason not to use score refinement.

\subsection{Late-fusion ensemble method}

We can apply this score refinement to tracklets of two different modalities rather than tracklets and detections of two consecutive time steps. Hence, we also implemented our approach as a late-fusion ensemble method. Whenever two modalities detect the same object, the confidence of its tracklet rises; otherwise, it drops. For comparison, other ensembles methods use voting strategies: the affirmative strategy takes all proposals, the consensus strategy takes those recognized by the majority, the unanimous strategy takes those identified by all modalities \cite{casado2020ensemble}. Since we only have two modalities, the consensus and unanimous strategies are the same here. 

We tested our implementation and compared it to the voting ensemble methods. As a baseline, we used our score refined CenterPoint as a LIDAR-based tracker and a corrected version of CenterTrack \cite{zhou2020tracking} as a camera-based tracker. CenterTrack appears to produce an error in its tracking ids. Remanaging its ids (foremost ensuring the id's uniqueness) turned out to have a significant effect on its performance, achieving an AMOTA of 17.75 on nuScenes validation set, in contrast to its reported AMOTA of 6.8 \cite{zhou2020tracking}, making it, in fact, one of the best camera-based trackers on the nuScenes data set by this day. In our previous experiments, we applied a score decay only on tracklets, not on new detections. Whereas with the fusion of two modalities, it seems more reasonable to apply score decay equally. Thus, we tested by employing a \textit{score-decay} of 0.2 to either of the modalities or both. 

In table \ref{tab:fusion}, we see the improvement of confidence-based ensemble methods with equal treatment of both modalities against other ensemble methods. We achieve the most significant improvement of 1.67 in AMOTA score against the LIDAR-based tracker if we treat both modalities equally and only apply a \textit{score-decay} if both modalities disagree. The other ensemble methods did not improve our baseline. While these strategies count the number of agreements between modalities, they do not account for confidence differences. We compared our late fusion ensemble method to the state-of-the-art trackers on nuScenes test evaluation, and table \ref{tab:leaderboard} shows the results. We see that our approach is not far away from the best performing method and comparable to the best-published work to date.



\section{Conclusion}
This work showed that \textit{confidence-based} MOT works better than \textit{count-based} MOT when using \textit{score-decay} and a proper score update function. We made an ablation study to see the effects of the hyperparameters and score update functions. We found out that most of the hyperparameters can be neglected but the score-decay, and multiplying the score's complements is the best score update function. We achieved a consistent improvement when using different detectors, filtering algorithms and datasets. We achieved an improvement up to 1.83 in AMOTA and up to 2.96 in MOTA scores. We used our approach as a late-fusion ensemble method in a multi-modal pipeline. It reinforced the score of matched tracklets from different modalities, resulting in an additional improvement of 1.67 in AMOTA. With a total AMOTA score of 67.6 on the nuScenes test set, we can show that our method is comparable to state-of-the-art methods. We believe that our approach is general and can be applied to even further tasks, such as 2D MOT.

\addtolength{\textheight}{-15cm}

\bibliographystyle{IEEEtran}
\bibliography{IEEEabrv,egbib}

\end{document}